\documentclass[sigconf]{acmart}

\usepackage{caption}
\usepackage[inline]{enumitem}
\usepackage{listings}
\usepackage{multirow}   
\usepackage{natbib}     
\usepackage{subcaption} 

\definecolor{greenish}{cmyk}{0.9,0,0.55,0}
\definecolor{purpleish}{RGB}{125,104,155}

\newcommand{\journalversion}[1]{}

\newcommand{\submittedorcameraready}[2]{#2}

\copyrightyear{2024}
\acmYear{2024}
\setcopyright{rightsretained}
\acmConference[UKICER 2024]{The United Kingdom and Ireland Computing Education Research}{September 5--6, 2024}{Manchester, United Kingdom}
\acmBooktitle{The United Kingdom and Ireland Computing Education Research (UKICER 2024), September 5--6, 2024, Manchester, United Kingdom}\acmDOI{10.1145/3689535.3689554}
\acmISBN{979-8-4007-1177-0/24/09}

\begin{document}

\title{Not the Silver Bullet}
\subtitle{LLM-enhanced Programming Error Messages are Ineffective in Practice}

\author{Eddie Antonio Santos}
\orcid{0000-0001-5337-715X}
\affiliation{%
  \institution{School of Computer Science}
  \institution{University College Dublin}
  \city{Dublin}
  \country{Ireland}
}
\email{eddie.santos@ucdconnect.ie}

\author{Brett A. Becker}
\orcid{0000-0003-1446-647X}
\affiliation{
  \institution{School of Computer Science}
  \institution{University College Dublin}
  \city{Dublin}
  \country{Ireland}
}
\email{brett.becker@ucd.ie}
\renewcommand{\shortauthors}{Santos and Becker}

%

\begin{abstract}
    The sudden emergence of large language models (LLMs) such as ChatGPT has had a disruptive impact throughout the computing education community.
    LLMs have been shown to excel at producing correct code to CS1 and CS2 problems, and can even act as friendly assistants to students learning how to code.
    Recent work shows that LLMs demonstrate unequivocally superior results in being able to explain and resolve compiler error messages---for decades,
    one of the most frustrating parts of learning how to code.
    However, LLM-generated error message explanations have only been assessed by expert programmers in artificial conditions.
    This work sought to understand how novice programmers resolve programming error messages (PEMs) in a more realistic scenario.
    We ran a within-subjects study with
    \textbf{$n = 106$} participants in which
    students were tasked to fix six buggy C programs. For each program,
    participants were randomly assigned to fix the problem using either a
    stock compiler error message, an expert-handwritten error message, or an
    error message explanation generated by GPT-4.
    Despite promising evidence on synthetic benchmarks,
    we found that GPT-4 generated error messages outperformed conventional compiler error messages in only 1 of the 6 tasks, measured by students' time-to-fix each problem.
    Handwritten explanations still outperform LLM and conventional error messages, both on objective and subjective measures.
\end{abstract}



\maketitle


\section{Introduction}

For decades, students learning how to code have struggled with error messages~\cite{becker2019compiler}---%
whether they are emitted by compilers or runtime systems,
programming error messages (PEMs) have had a reputation for being
terse~\cite{barron1975note},
inadequate~\cite{brown1983error},
and unreadable~\cite{denny2021designing}.
Error messages from C and C++ compilers especially have been shown to be deficient debugging tools~\cite{traver2010compiler}.
\journalversion{
    Difficulty with PEMs also contributes to the notion that ``programming is hard'' which has many negative consequences~\cite{becker2021what}.
}

Recent advances in generative AI have resulted in tools like ChatGPT and GitHub Copilot.
These tools, based on large language models (LLMs), have revolutionised several fields including computing education~\cite{becker2023programming}.
Recent work has shown that LLMs can produce acceptable programming error message explanations~\cite{leinonen2023using} which become more accurate with larger models and more source code context~\cite{widjojo2023addressing,santos2023always}.
However, it is unknown to what extent that novice programmers are able to effectively utilise these automatically generated error message explanations to debug their programs.

In this study, we had 106 students from an introductory programming module partake in a within-subjects study that had participants fix a number of buggy programs.
For each program, participants were shown either a conventional compiler error message, an expert-handwritten error message, or an LLM-generated explanation.
The LLM used to enhance error messages was GPT-4, which at the time of the study, was the LLM used in the paid version of ChatGPT~\cite{openai2023gpt4}.\@
We measured both how long it took participants to resolve errors with each condition, as well as asking students their opinions on their debugging experience.

\subsection{Contributions \& Research Questions}

We provide empirical evidence demonstrating that students appear not to be any faster at resolving errors when given GPT-4 error message explanations compared to stock compiler error messages. Even though students are not any faster, they still prefer GPT-4's explanations to conventional compiler error messages. However, expert-handwritten are superior to both.
We posit that error message usability is more complex than whether or not the text presented to the student contains the correct solution for the problem.

The following questions guide this research:
\begin{enumerate}
    \item[RQ1] How quickly do students resolve error messages when given LLM-enhanced explanations in comparison to stock compiler error messages and handwritten explanations?
    \item[RQ2] Which style of error message do students prefer?
\end{enumerate}

\section{Background and Related Work}

\subsection{Programming Error Messages}


Programming error messages (PEMs) are the diagnostic messages presented to coders when an error is detected in their program---%
either due to a mistake in syntax or spelling in the source code,
or due to some unrecoverable runtime condition, such as a division by zero, or an invalid memory access~\cite{becker2019compiler}.
PEMs have been an obstacle to learning how to code since almost the inception of programming~\cite{barron1975note,wexelblat1976maxims,brown1983error}.
Little progress has been made to improve compiler and runtime error messages, despite
decades of guidelines proposed to improve them~\cite{horning1974what,traver2010compiler,barik2018error,kohn2019error,meredith2019writing,denny2021designing}.
There have been many attempts at having programming systems produce better diagnostics~\cite{rosen1965pufft,moulton1967ditran,burgess1972compile,schorsch1995cap,kohn2020tell},
however, error message enhancement has seen weak~\cite{becker2016effective,becker2018effects,harker2017examining,prather2017novices} to insignificant~\cite{denny2014enhancing,pettit2017enhanced} results in improving student outcomes.
\journalversion{%
Error messages have plagued programming for so long that modern programming languages and environments often boast about their ``good'' error messages~\cite{rustc2024errors,czaplicki2015compilers,llvm2009clang}.%
}

\subsection{Large Language Models and CS Education}%
\label{sec:llms-csed}

A confluence of advances in model architecture;
novel text representation;
massive, curated datasets;
and sheer computing power has rapidly enabled the development of large language models (LLMs)~\cite{ramponi2023full}:
models with billions or even trillions of parameters,
capable of capturing the structure and predictability of text to such an extent that they are able to exhibit
``emergent'' behaviours, like question answering,
analogical reasoning,
and even the ability to execute programs~\cite{wei2022emergent}.

In computing education, LLM-powered tools have been shown to ace 
CS1~\cite{finnieansley2022robots} and CS2~\cite{finnieansley2023myai} exams
and provide increasingly accurate error message explanations~\cite{leinonen2023using,santos2023always,widjojo2023addressing}.
LLMs have even enabled brand new pedagogical approaches~\cite{nguyen2024beginning,denny2024prompt}.
Educators are grappling with how to integrate LLMs into their practice~\cite{becker2023programming}---if at all~\cite{lau2023ban}.
Without guidance, complete novices struggle to write the prompts that would complete their assignments~\cite{nguyen2024beginning}.
Additionally, novices exhibit a number of unproductive interaction patterns when using LLM-assisted code completion~\cite{prather2023weird,vaithilingam2022expectation}.
Some programmers do not complete programming tasks faster with LLM-assisted code completion, and in fact, are more likely to fail programming tasks~\cite{vaithilingam2022expectation}.
Having an LLM that performs better in synthetic benchmarks results in ``relatively indistinguishable differences in terms of human performance''~\cite{mozannar2024realhumaneval}.
Despite the lack of improvement, novices express a preference for using LLMs and chatbots~\cite{cipriano2024chatgpt,prather2023weird,vaithilingam2022expectation}.
However, more experienced students express concern with how LLMs may hamper their learning~\cite{prasad2023generating}.

\paragraph{LLMs and PEMs}
LLMs have been found to be useful at explaining programming error messages on synthetic benchmarks.
\citet{leinonen2023using} used OpenAI Codex to explain Python error messages.
They found that the best, most accurate explanations and fixes were obtained when providing source code in the prompt,
as well as using a temperature value of 0 (explained in \autoref{sec:gpt4}).
Their prompt forms the basis of the prompt that was used in our study.
\citet{santos2023always} and \citet{widjojo2023addressing} have similar findings:
providing Java source code in the prompt produces significantly better error message explanations.
Additionally, more advanced models like GPT-4 are more likely to output accurate explanations and fixes than GPT-3 and Codex.

\section{Methodology}

We conducted a within-subjects study, inspired by prior work~\cite{vaithilingam2022expectation}.
Each participant observed all three study conditions---control, handwritten, and GPT-4 (Section~\ref{sec:conditions}).
Each participant was tasked to fix all six buggy C programs (Section~\ref{sec:tasks}).
Both condition assignment and task assignment were randomised to counterbalance responses,
such that we would obtain a roughly equal amount of responses for each task/condition pair.
Randomising participants' assignments also helped to mitigate the learning effect. 
Having participants fix bugs under all three study conditions allowed them to directly compare the different error message styles to one another, and report which style they preferred.
The study began with a short questionnaire, after which participants were given an in-person briefing, which was followed by the six debugging tasks.
After participants had completed all six tasks, we asked participants questions to compare the three error message styles directly.
The remainder of this section describes the study conditions, the tasks, and study protocol in greater detail.

\subsection{Study conditions}%
\label{sec:conditions}

\begin{figure}[tbh]
    \centering
    \begin{subfigure}[b]{\columnwidth}
        \includegraphics[width=\columnwidth]{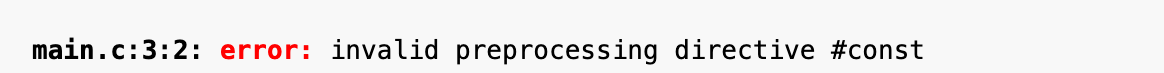}
        \Description{main.c:3:2: error: invalid preprocessing directive #const}
        \caption{GCC (control)}\label{subfig:gcc-pem}
    \end{subfigure}
    \begin{subfigure}[b]{\columnwidth}
        \includegraphics[width=\columnwidth]{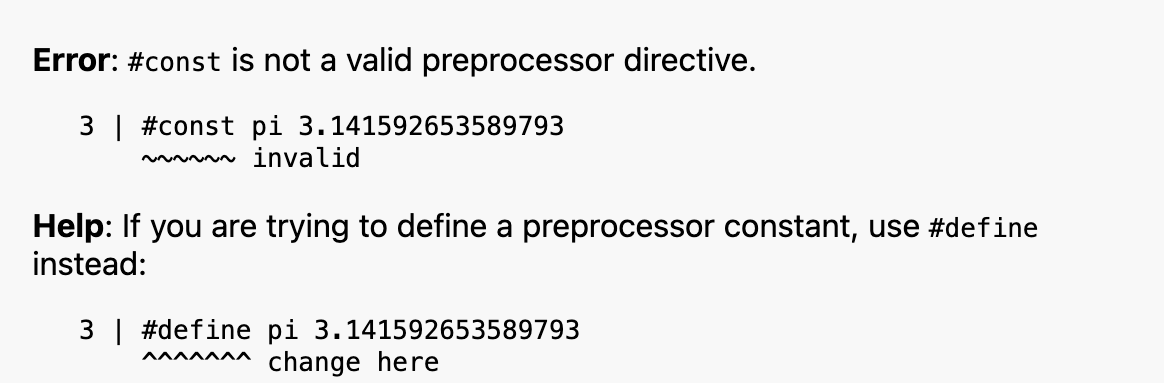}%
        \Description{Error: #const is not a valid preprocessor directive.  Help: If you are trying to define a preprocessor constant, use  #define
  instead: [code is highlighted to show where #define would be used instead]}
        \caption{Handwritten}\label{subfig:handwritten-pem}
    \end{subfigure}
    \begin{subfigure}[b]{\columnwidth}
        \includegraphics[width=\columnwidth]{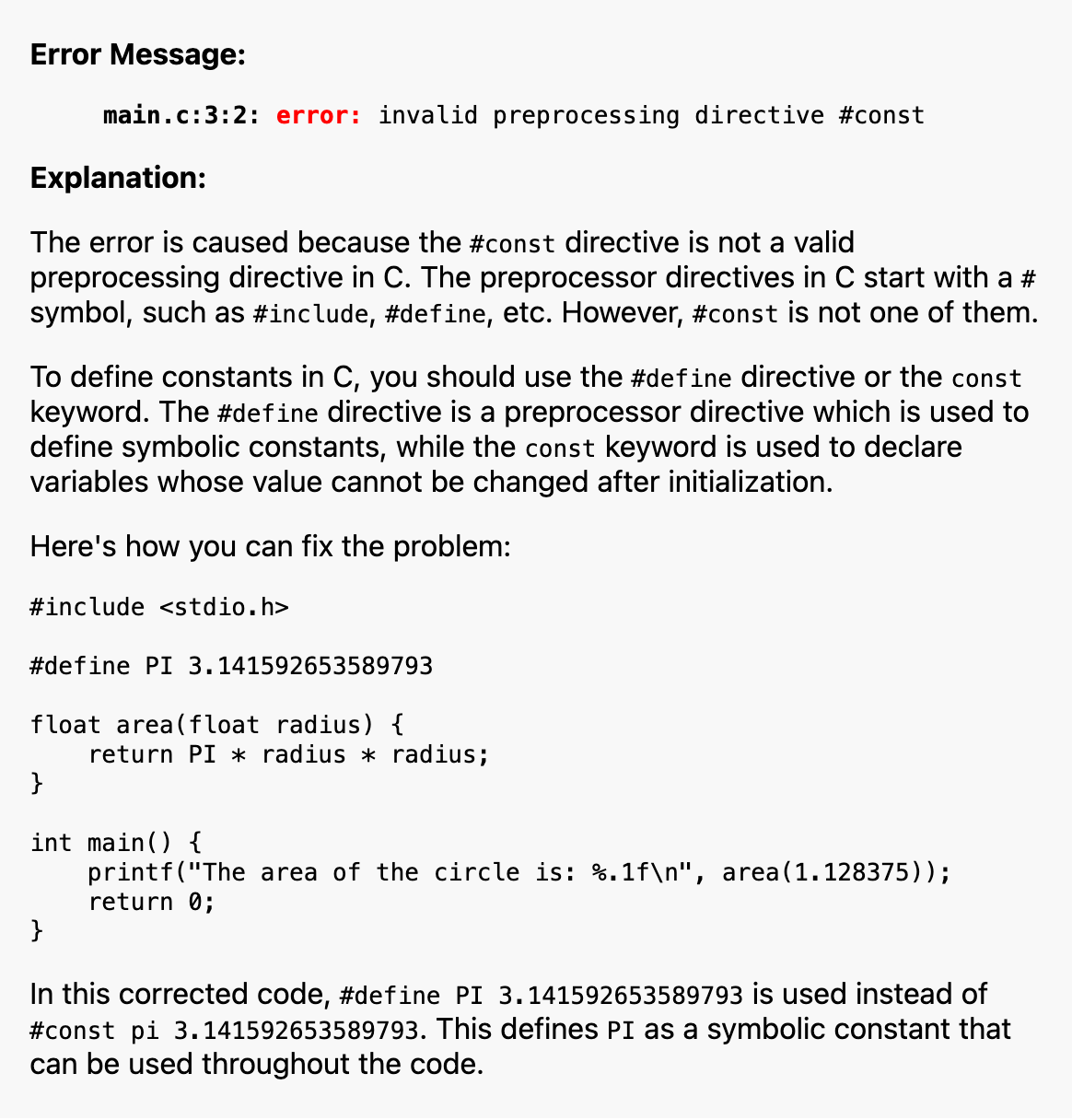}%
        \Description{The error is caused because the  #const  directive is not a valid preprocessing directive in C. The preprocessor directives in C start with a #  symbol, such as  #include ,  #define , etc. However,  #const  is not one of them.

        To define constants in C, you should use the  #define  directive or the const  keyword. The  #define  directive is a preprocessor directive which is used to define symbolic constants, while the  const  keyword is used to declare variables whose value cannot be changed after initialization.

        Here's how you can fix the problem: [the full source code is reproduced with #define instead of #const]

        In this corrected code,  #define PI 3.141592653589793  is used instead of #const pi 3.141592653589793 . This defines  PI  as a symbolic constant that can be used throughout the code.
        }
        \caption{GPT-4}\label{subfig:gpt4-pem}
    \end{subfigure}
    \caption{Examples of the three error message styles (i.e., study conditions) as they would appear to participants.}%
    \label{fig:conditions}
\end{figure}

Each participant saw error messages presented in all three study conditions: \textbf{GCC} (control), \textbf{handwritten}, and \textbf{GPT-4}.
However, the order of the study conditions was shuffled for each student,
so that participants could directly compare the three study conditions themselves.
Since there were six tasks, but only three conditions,
the order of the study conditions was simply repeated for the latter three tasks---%
for example,
if a participant was assigned handwritten for the first task, GCC for the second, and GPT-4 for the third,
then they would be assigned handwritten for the fourth, GCC for the fifth, and GPT-4 for the sixth.

\paragraph{Control: GCC}
For the control condition, students were presented with error messages directly obtained from the GCC 13.2.0 C compiler (\autoref{subfig:gcc-pem}).
Whenever a single programming error would induce multiple spurious, cascading error messages,
we would only show the first error message, as advised by Becker et al.~\cite{becker2018fix}.

\paragraph{Handwritten error messages}
Error message explanations (\autoref{subfig:handwritten-pem}) were handwritten by the first author.
These explanations were written in response to the problems present in the source code,
but not necessarily in response to any error messages emitted by GCC.\@
Importantly, the handwritten explanations were finalised \emph{before} error message explanations were obtained from GPT-4.
Therefore, the author of the handwritten explanations was not influenced by GPT-4's output.
Every message was written in a consistent structure:
first was a line beginning with the word \textbf{Error:}\ which states what the detected error is,
followed by a relevant excerpt from the source code.
Then one or more sections beginning with the word \textbf{Help:}\ or \textbf{Note:}\ would either suggest a possible solution, or highlight relevant information to fix the problem.
The structure of the messages was greatly inspired by the diagnostics emitted by the Rust compiler,
with source code excerpts mimicking the structure of Rust's ``diagnostic windows''~\cite{rustc2024errors}.
The handwritten explanations were written in such a way that they can plausibly be generated by an actual compiler,
given sufficient context.

\paragraph{GPT-4 enhanced error messages}%
\label{sec:gpt4}
After the handwritten error explanations were written,
we obtained error message explanations using OpenAI's GPT-4 API.\@
All GPT-4 responses were obtained before the study started, as to not include inference time in the time-to-fix measure.\footnote{
    Obtaining the full output for each of these prompts would take between 16--30 seconds, depending on the problem.
}
The methodology for generating error message explanations is derived from prior work~\cite{leinonen2023using,santos2023always}.
On January 25, 2024, we used the OpenAI API to prompt GPT-4 model \texttt{gpt-4-0613}.
Each prompt used the system message of ``You are a helpful assistant''.
We prompted with a temperature of 0, a hyperparameter used to affect the determinism of an LLM's output, where 0 indicates the most deterministic and reproducible output.\footnote{%
    A value greater than 0 introduces entropy into the sampling distributions, in a style reminiscent of Ludwig Boltzmann's work on thermodynamics.
}
The prompt used was identical to that in \citet{santos2023always}, providing both the error message verbatim from GCC as well as the full source code of the buggy program.
%
%

When presenting the error message to participants (\autoref{subfig:gpt4-pem}),
the message would start with \textbf{Error Message:}\ followed by the GCC error message.
This is because GPT-4's output would often make reference to the original error message in its explanations.
After this, a line starting with \textbf{Explanation:}\ would be followed by GPT-4's output, rendered as Markdown.
GPT-4 output would vary depending on the problem,
however, every response consistently had a line saying something similar to ``here's the corrected code:''
followed by a full reproduction of the source code with the problem resolved.
In all six problems, GPT-4 generated a solution equivalent to the one suggested in the handwritten explanations.
In addition, GPT-4's error message explanations were devoid of any major technical inaccuracies or ``hallucinations''~\cite{huang2023survey}.

\subsection{Tasks}%
\label{sec:tasks}

Students were to fix all of the following C programming errors, in a randomly assigned order.
All programs in this study caused GCC to emit compiler error messages.
We did not have students debug problems that would result in runtime errors
(e.g., no segmentation faults).
The following are all six of the debugging tasks:

\begin{enumerate}
    \item\textbf{Flipped assignment}.
    The left- and right-hand sides of an assignment statement were swapped
    such that the assignment target would be on the right-hand side, e.g.,
    \lstinline{a + b = c}.

    \item\textbf{\lstinline{\#const} instead of \lstinline{\#define}}.
    A program was made attempting to define a constant called \lstinline{PI} using the syntax \lstinline{#const PI} \lstinline{3.14}.
    This programming error intentionally conflates C's \lstinline{#define} preprocessor directive with the \lstinline{const} type qualifier, both being valid ways to define a constant in C.

    \item\textbf{Using a keyword as a name}.
    The program attempts to create new variables called \lstinline{union} and \lstinline{nonUnion}, however, \lstinline{union} is a reserved word, and cannot be used as an identifier. 

    \item\textbf{Missing parameter}.
    A function was defined to convert from Fahrenheit to Celsius, however, the function definition lacks any formal parameters.
    Despite this, the body of the function used a identifier \lstinline{fahrenheit} to perform the conversion, and the function is called with an argument for the temperature in Fahrenheit.

    \item\textbf{Missing curly brace}.
    The opening curly brace of a function definition was omitted, causing GCC's parser to take a ``garden-path'' and completely misinterpret the program.

    \item\textbf{Reassigning a constant}.
    A formal parameter was declared \lstinline{const}, then reassigned within the function body.
\end{enumerate}

\subsection{Participants}%
\label{sec:participants}

Participants were recruited from a class at a large research-intensive European public university that would be an R1 in the US Carnegie Classification.
The class was the second semester of the first-year programming sequence (CS1) for CS majors, taught in the C programming language.
In total, 113 participants were recruited across two separate lab sections.
Of those, \(n = 106\) participants (94\%) completed the study.
Of the participants that completed the study, 78 identified as men, 23 as women, and 5 chose not to disclose their gender.
At the beginning of the study, we asked participants a few questions on their experience in programming.
The most commonly reported experience level was between 0--3 months (32 participants).
Curiously, 8 participants reported absolutely zero experience in programming, despite being enrolled in the second semester of the programming sequence.
The remaining 66 participants reported greater than three months of programming experience.

\subsection{Protocol}%
\label{sec:protocol}

\begin{figure}[tbh]
    \centering
    \includegraphics[width=\columnwidth]{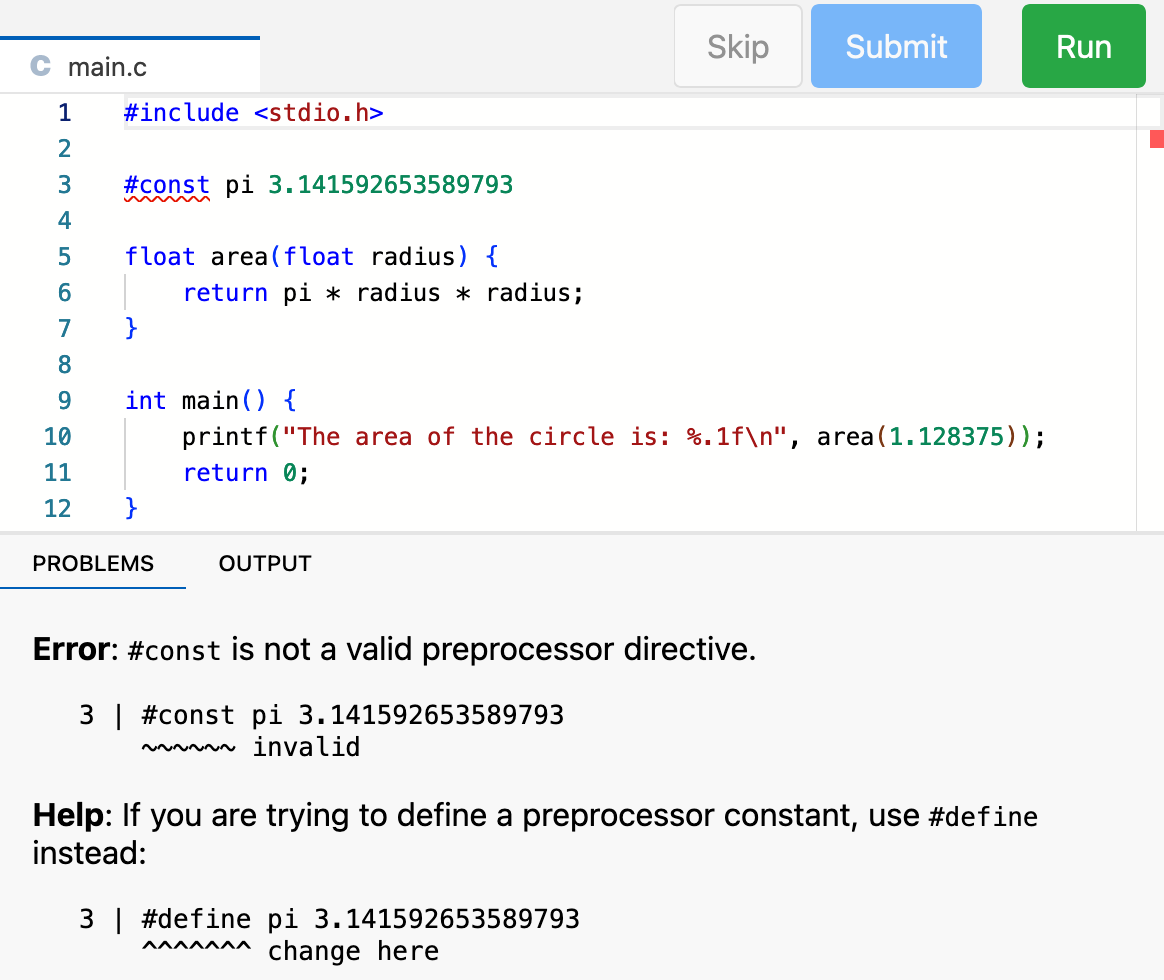}
    \caption{Screenshot of the web-based IDE, showing the ``\mbox{\lstinline{\#const}} instead of \lstinline{\#define}'' task under the handwritten error message condition.}%
    \label{fig:ide-screenie}
    \Description{
        Screenshot of an IDE, much like a stripped-down version of Visual Studio Code. At the top right, there are the buttons ``Skip'', ``Submit'', and ``Run'', with only both ``skip'' and ``submit'' being dimmed.
        In the middle is a code pane with a C program with a mistake in it.
        At the bottom, there is a ``problems'' pane, which displays the handwritten error message in the above figure.
    }
\end{figure}

The study was conducted during a regularly scheduled lab session in late January 2024---%
the second scheduled lab of the semester---%
held simultaneously in two separate classrooms.
Students were under no obligation to participate in the lab session;
they were not compensated for participation,
and participation did not affect the students' grade in any way.

Once in the classroom, the study was conducted entirely via a custom web-based survey platform,
which combined multiple questionnaires with an online IDE (\autoref{fig:ide-screenie}).
The online IDE component was created using Microsoft's Monaco Editor,
the core text editing component of Visual Studio Code~\cite{microsoft2024monaco}.
Upon hitting the ``Run'' button in the IDE, the full source code was securely transmitted to a university-managed server, where it would be stored.
Code was compiled and run in a sandboxed environment provided by the Piston code execution engine~\cite{seymour2023piston}.

At the start of the study, the first author was present and gave a quick oral briefing.
After the briefing, students were directed to the web platform, where they signed an online consent form to commence the study.
First, participants were asked a few demographic questions and asked about their attitudes regarding programming and error messages.
After this, participants started the six debugging tasks.

For each task, students were presented with the buggy code in the web IDE (\autoref{fig:ide-screenie}) and were instructed to fix the problem,
hitting the ``Run'' button and only submitting their solution once the problem was resolved.
We started measuring the time taken to complete the task from the moment the code editor loaded.
After 5 minutes, students were given the option to skip the current task (if they were stuck on a problem and wanted to continue to the next one).
Students were given up to 10 minutes per task, after which their attempt would have timed-out;
however, all 106 participants either submitted or skipped each task before the 10 minute time-out would have taken effect.
During the study, postgraduate TAs were instructed to ensure that students were not using AI tools like ChatGPT/GitHub Copilot.
However, the students were still permitted to search the web for error message explanations and use Q\&A sites like Stack Overflow.
After each exercise, students were given a short questionnaire to reflect on how they used the error message to solve (or not solve) the programming error.
This was presented as four Likert-type questions (\autoref{fig:likert}), followed by a question asking about the message's length. 
In total, 106 participants completed all six tasks within 50 minutes.
%

\section{Results}

\subsection{Objective measures}%
\label{sec:objective}

The primary quantitative measurements that we recorded were the time-to-fix for participants who successfully fixed a task,
and whether or not a student skipped a particular task.
For each task/condition pair, we obtained between between 27--44 samples, which is sufficient to perform within-task comparisons.

%
%
%

\newcommand{\faster}[1]{\textbf{#1}}
\begin{table}[tbh]
\centering
\caption{Tukey's HSD test of the means of log-transformed time-to-fix, comparing the error message condition. The difference in median time-to-fix between the left and right condition is given. \faster{Bold font} indicates the condition with the faster (statistically significant) time-to-fix.}%
\label{tab:comparisons}
\footnotesize
\begin{tabular}{@{}lllrr@{}}
\toprule
Task                                        & \multicolumn{2}{c}{Comparison}                        & Diff                         & \(p\)            \\ \midrule
\#const instead of \#define                 & GCC                            & \faster{Handwritten} & 43.95s                       & 0.003            \\
                                            & GCC                            & GPT-4                & 4.93s                        & 0.893            \\
                                            & \faster{Handwritten}           & GPT-4                & -39.02s                      & 0.002            \\
                            \rule{0pt}{4ex}
Using a keyword as a name                   & GCC                            & \faster{Handwritten} & 63.42s                       & \textless{}0.001 \\
                                            & GCC                            & \faster{GPT-4}       & 28.61s                       & 0.006            \\
                                            & \faster{Handwritten}           & GPT-4                & -34.82s                      & \textless{}0.001 \\
                            \rule{0pt}{4ex}
Missing parameter                           & GCC                            & Handwritten          &  0.54s                       & 0.716            \\
                                            & \faster{GCC}                   & GPT-4                & -15.08s                      & 0.021            \\
                                            & Handwritten                    & GPT-4                & -15.61s                      & 0.157            \\
                            \rule{0pt}{4ex}
Missing curly brace                         & GCC                            & \faster{Handwritten} & 122.3s                       & \textless{}0.001 \\
                                            & GCC                            & GPT-4                & 50.6s                        & 0.373            \\
                                            & \faster{Handwritten}           & GPT-4                & -71.6s                       & \textless{}0.001 \\
                            \rule{0pt}{4ex}
Reassigning a constant                      & GCC                            & \faster{Handwritten} & 53.0s                        & 0.031            \\
                                            & GCC                            & GPT-4                & 6.28s                        & 0.996            \\
                                            & \faster{Handwritten}           & GPT-4                & -51.6s                       & 0.029            \\ \bottomrule

\end{tabular}

\end{table}

\begin{figure}[tbh]
    \centering
    \includegraphics[width=\columnwidth]{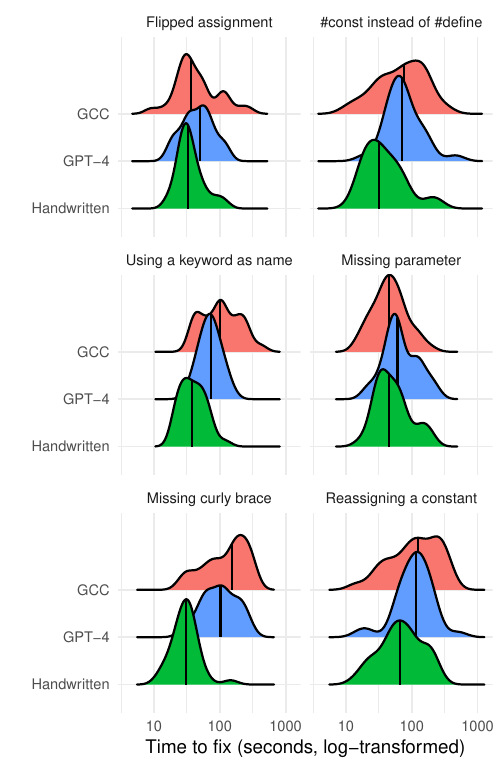}
    \caption{Density  plots of the log-transformed time-to-fix, by task. The black vertical line denotes the median.}%
    \label{fig:per-task-density-log}
    \Description{
        A series of stacked density plots, one per each of the six tasks.
    }
\end{figure}

\paragraph{Time-to-fix}%
\label{sec:time-to-fix}
Time-to-fix is consistently right-tailed, so we log-transformed the data to perform statistical comparisons.
We noticed that each task given had an effect on time-to-fix,
so we performed within-task comparisons.
For each task, we performed a one-way ANOVA to compare the effect of the study condition on students' log-transformed time-to-fix (\autoref{fig:per-task-density-log}).
A one-way ANOVA revealed that there was a statistically significant ($p < 0.05$) difference in the mean log-transformed time-to-fix between the conditions in all tasks,
except for the ``flipped assignment'' task, in which no statistically significant difference was found ($p = 0.147$).
We will omit the flipped assignment task for the remainder of this section.
For the five remaining tasks, we performed Tukey's HSD test
on the means of log-transformed time-to-fix (\autoref{tab:comparisons}).
Because the mean of log-transformed time-to-fix is not useful to report, we instead report the difference in median time-to-fix, as median is preserved after log-transformation.
We found that GPT-4 outperforms the control (GCC error messages) in only one of the five remaining tasks, namely, ``using a keyword as a name''.
In three of the five tasks, we could not find a statistically significant difference in the mean log-transformed time-to-fix between GPT-4 and the control.
However, the handwritten error messages outperformed both the control and the GPT-4 enhanced error messages in all tasks, except the ``missing parameter'' task,
where we could not find a statistically significant difference between the handwritten error messages and the control;
however, stock GCC error messages outperformed GPT-4's explanations.

\paragraph{Skip rate}%
\label{sec:skip-rate}
We gave the option for students to skip exercises after spending 5 minutes on them without submitting a solution.
Overall, students rarely skipped exercises, with only 13 skips out of 636 exercise attempts (2.0\%).
In other words, students successfully fixed the programming errors in 98\% of all exercises.
There were zero skips observed for students fixing tasks under the handwritten condition.
The most skips (10) were observed for the control condition, whereas GPT-4 had only 3 skips.
A $\chi^2$-test failed to find a statistically significant difference between the three conditions.

\subsection{Subjective measures}%
\label{sec:subjective}

\submittedorcameraready{}{
    \begin{figure}[tbh]
        \centering
        \includegraphics[width=\columnwidth]{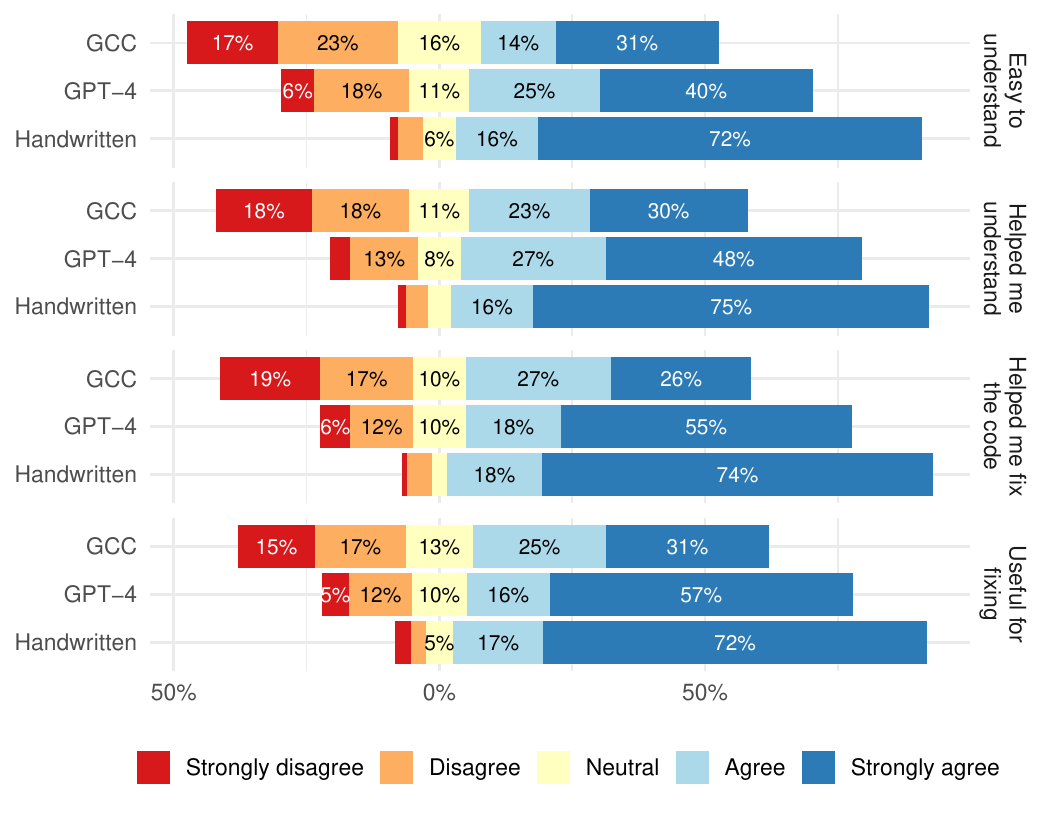}
        \caption{%
            Proportion of participants' Likert responses.
            From top-to-bottom, the questions were
            ``The message was easy to understand'',
            ``The message helped me understand what was wrong with the code'',
            ``The message helped me fix the code'',
            and
            ``The message was useful for fixing the problem''.%
        }%
        \label{fig:likert}
        \Description{%
            Likert responses for four questions.
            For all questions, the GCC (control) has the least favourable responses,
            while GPT-4 has the medium responses, and handwritten has by far the most
            favourable responses.%
        }
    \end{figure}
}

\paragraph{Opinion}
After each exercise, we asked participants four Likert-type questions to gauge
\submittedorcameraready{%
    their opinion on  two axes:
    \begin{enumerate*}
        \item how easy the error message was to understand; and
        \item how much participants felt that the error message helped them fix the code.
    \end{enumerate*}
    We performed a linear regression to predict participants' response to both axes given the study conditions.
    Using a handwritten message results in a 1.316 increase in positive opinion for how easy a message is to understand compared to the control (\(p < 0.01\));
    whereas using a GPT-4 message results in a 0.656 increase in opinion for how easy a message is to understand compared to the control (\(p < 0.01\), \(R^2 = 0.158\)).
    For the helpfulness axis,
    we found that using handwritten compared to the control will result in a 1.231 increase in opinion on how useful a message is for fixing the code (\(p < 0.01\)),
    whereas using GPT-4 will result in a 0.733 increase compared to control in opinion on how useful a message is for fixing the code (\(p < 0.01\), \(R^2 =  0.145\)).
}{%
    their general opinion on how useful the message was for fixing the problem (\autoref{fig:likert}).
    We performed a linear regression to predict participants' overall opinion given the study conditions.
    We modelled opinion as a numerical variable where
    \newcommand{\likertresponse}[1]{\textit{#1}}
    \likertresponse{Strongly disagree} = -2,
    \likertresponse{Disagree} = -1,
    \likertresponse{Neutral} = 0,
    \likertresponse{Agree} = 1,
    and \likertresponse{Strongly agree} = 2,
    then took the mean of a participant's responses per each condition.
    Using a handwritten message results in a 1.27 point increase in opinion compared to traditional compiler error messages (\(p < 0.001\));
    whereas using a GPT-4 generated message results in a 0.69 point increase in opinion compared to the control (\(p < 0.001\)).
}
Overall, we found that participants rated both the handwritten messages and GPT-4 explanations higher than GCC's error messages,
with the handwritten error messages being the most highly rated. 
Participants rated GPT-4 error messages highly in terms of being useful to help them solve the error,
despite both conditions suggesting equivalent fixes.

\submittedorcameraready{
    \begin{figure}[tbh]
        \centering
        \includegraphics[width=\columnwidth]{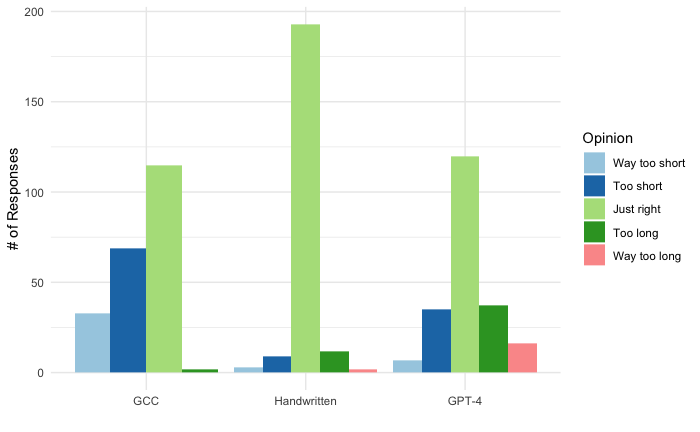}
        \caption{Participants' overall opinions on the length of error messages, per condition.}%
        \label{fig:message-lengths}
        \Description{
            Three histograms of students' opinions of message length.
        }
    \end{figure}
}{}

\paragraph{Message length}%
\label{sec:message-length}
\submittedorcameraready{%
    In addition to opinion, we also asked participants to rate the length of the error message after each exercise.
    \autoref{fig:message-lengths} shows participants' overall opinions of the length of error messages.
}{
    In addition to opinion, we also asked participants to rate the length of the error message after each exercise, on a scale from
    \textit{way too short},
    \textit{too short},
    \textit{just right},
    \textit{too long}, to
    \textit{way too long}.
}
Participants overwhelmingly found that the handwritten error messages were just the right length (88.1\% of responses).
GCC's error messages were mostly deemed as either the right length or too short---never way too long.
GPT-4's error messages were roughly binomially distributed across all five categories.

\section{Discussion}

Despite promising evidence in prior studies~\cite{widjojo2023addressing,santos2023always,leinonen2023using},
GPT-4 error message explanations do not help novices when they are resolving error messages as much as one would expect.
In fact, in one of the tasks (``missing parameter''), students were \emph{slower} when using GPT-4's explanation.
Curiously, expert-handwritten error messages outperform GPT-4 error message explanations even though both suggested equivalent solutions for each problem.

When it comes to students' preferences,
they preferred GPT-4's error messages over GCC's terse, jargon-heavy error messages.
This makes sense, as GPT-4 would always produced full, complete sentences---%
a factor that was previously found to be important to error message
readability~\cite{denny2021designing}.
We were surprised that participants did not report GPT-4's error messages as being too long.
That said, students were unable to use these messages effectively,
even though GPT-4's messages would always provide the correct way to solve the programming error.
Prior work has found that longer error messages do not seem to help students~\cite{nienaltowski2008compiler}.

\paragraph{Programming is (still) hard}
Early results in understanding LLMs' capabilities at introductory programming seemed promising~\cite{finnieansley2022robots,finnieansley2023myai,santos2023always,leinonen2023using},
inspiring researchers to declare a new era for computing education~\cite{becker2023programming} and even ``the end of programming''~\cite{welsh2022end}.
However, it seems that LLMs are not the transformative tool that they once seemed.
Our findings---that, despite excelling in synthetic benchmarks, LLMs do not significantly improve programmers' productivity---%
are corroborated by a number of studies~\cite{vaithilingam2022expectation,nguyen2024beginning,mozannar2024realhumaneval,howardsarin2024future}.
Additionally, participants express preference for LLMs~\cite{prather2023weird,vaithilingam2022expectation},
even though LLMs' answers do not make them more effective at resolving programming errors.
\citet{simkute2024ironies} argue that this supposed contradiction of LLM productivity is an already well-known phenomenon in human factors research: automation alters peoples' workflows in unproductive ways, such as turning active producers into passive evaluators.
This change in workflow widens the gap between the programmer's mental model,
and what the programmer \emph{ought} to attend to while solving problems.
In fact, there is evidence that LLM-powered code suggestion alters programmers' workflows in ways that hamper productivity~\cite{prather2023weird,mozannar2022reading}.
LLMs have not delivered on the promise of natural language programming~\cite{nguyen2024beginning};
rather, they provide an indirect method of manipulating an existing abstraction: high-level source code.
Similarly, LLMs have not fundamentally altered the task of debugging;
they just explain the already difficult problem in a more approachable manner.
Thus, debugging remains just as difficult as it was prior to the introduction of LLMs.

\journalversion{
    \textcolor{red}{TODO: programming is not hard, our expectations are too high.
    LLMs have not changed debugging, so implore educators to focus on teaching debugging.
    LLMs can explain errors in a more approachable manner, but care must be
    taken in helping students use LLMs' answers effectively.}
}

\subsection{Limitations}%
\label{sec:limitations}

The generalisability of these results is limited by the sample of participants: all from one class at a European research university, taught in one programming environment.
Additionally, since the debugging tasks were created for the purpose of this study, the programming errors may be inauthentic to the kind of debugging that would naturally occur during programming.
Another limitation was that the same person who prepared the debugging tasks also wrote the handwritten error messages.
There is also the confound in that the GPT-4 error messages always presented the stock compiler error message verbatim, rather than produce their own explanation of the error, without using the original error message as context.
It is possible that participants read the stock compiler error message, without reading further to use the GPT-4 explanation.

\section{Conclusion}

We conducted a within-subjects experiment where novice programmers fixed six buggy programs using three different error message styles.
In contrast to results on synthetic benchmarks, GPT-4 enhanced error messages were only more effective than stock compiler error messages in one of the six programming tasks.
Handwritten error messages were more effective than the control in four of six tasks, and more effective than GPT-4's error message explanations in five of six tasks.
Overall, students preferred GPT-4's messages over stock compiler error messages, but preferred the handwritten error messages even more.
This is despite the fact that the GPT-4 error message explanations and the handwritten error messages both made the same suggestion to fix the problem.
Future work should further understand what factors make an error message usable and how programming environments and LLMs alike can be modified to satisfy novices' needs.
\submittedorcameraready{
    It appears that we have a long way to go to reach ``the end of programming''.
}{
    It appears we still have a long way to go to reach ``the end of programming''.
}

\begin{acks}
Thanks to Dennis Bouvier for inspiration on creating the debugging tasks;
Gavin McArdle and Di Meng for graciously helping us run the study in their lab sessions;
Simon Caton, Kira Finan, and Olivia Finan for help with data analysis;
and Sajjad Karimian, Fionn Murphy, and Abdul Wadud for beta testing the online IDE.
This study received approval from our institutional ethics review board (LS-23-56-Santos-Becker).
\end{acks}

\bibliographystyle{ACM-Reference-Format}
\bibliography{references}

\end{document}